*Research Article*

# Deep Neural Watermarking for Robust Copyright Protection in 3D Point Clouds

**Khandoker Ashik Uz Zaman[1], Mohammad Zahangir Alam[2], Mohammed N. M. Ali[1] and Mahdi H. Miraz[1,3,4,*]**

[1]Xiamen University Malaya, Malaysia
mcs2509008@xmu.edu.my; m.miraz@ieee.org; ali.mohammed@xmu.edu.my

[2]Brunel University of London
mohammad.alam@brunel.ac.uk

[3]Wrexham University, UK
m.miraz@ieee.org

[4]University of South Wales, UK
m.miraz@ieee.org

*Correspondence: m.miraz@ieee.org

**Abstract:** The protection of intellectual property has become critical due to the rapid growth of three-dimensional content in digital media. Unlike traditional images or videos, 3D point clouds present unique challenges for copyright enforcement, as they are especially vulnerable to a range of geometric and non-geometric attacks that can easily degrade or remove conventional watermark signals. In this paper, we address these challenges by proposing a robust deep neural watermarking framework for 3D point cloud copyright protection and ownership verification. Our approach embeds binary watermarks into the singular values of 3D point cloud blocks using spectral decomposition, i.e. Singular Value Decomposition (SVD), and leverages the extraction capabilities of Deep Learning using PointNet++ neural network architecture. The network is trained to reliably extract watermarks even after the data undergoes various attacks such as rotation, scaling, noise, cropping and signal distortions. We validated our method using the publicly available ModelNet40 dataset, demonstrating that deep learning-based extraction significantly outperforms traditional SVD-based techniques under challenging conditions. Our experimental evaluation demonstrates that the deep learning-based extraction approach significantly outperforms existing SVD-based methods with deep learning achieving bitwise accuracy up to 0.83 and Intersection over Union (IoU) of 0.80, compared to SVD achieving a bitwise accuracy of 0.58 and IoU of 0.26 for the Crop (70%) attack, which is the most severe geometric distortion in our experiment. This demonstrates our method's ability to achieve superior watermark recovery and maintain high fidelity even under severe distortions. Through the integration of conventional spectral methods and modern neural architectures, our hybrid approach establishes a new standard for robust and reliable copyright protection in 3D digital environments. Our work provides a promising approach to intellectual property protection in the growing 3D media sector, meeting crucial demands in gaming, virtual reality, medical imaging and digital content creation.

**Keywords:** *3D point cloud; copyright protection; digital watermarking; deep learning; robust watermark extraction; singular value decomposition; spectral decomposition.*

## 1. Introduction

The swift rise in the popularity of three-dimensional (3D) digital content and improvements in 3D scanning technology have completely changed how digital media works, enabling new possibilities in fields such as computer graphics, gaming, virtual and augmented reality, autonomous vehicles, medical imaging and more [1-4]. Amongst various 3D representations, point clouds have emerged as a widely adopted format owing to their simplicity and versatility in capturing geometric structures without reliance on explicit connectivity information [5-6]. As 3D point clouds become increasingly common in both commercial and research settings, concerns regarding unauthorised use, replication and distribution have intensified, highlighting the urgent need for robust copyright protection and ownership-verication mechanisms [7].

However, effective intellectual property protection for 3D point clouds presents unique challenges compared to traditional 2D images or videos. Point clouds are inherently unordered, lack a regular grid structure and are highly susceptible to a variety of geometric and non-geometric transformations [8]. Conventional digital watermarking techniques, which have proven effective for images and videos, often fail when applied to point clouds, as embedded signals can be easily degraded or erased by attacks such as rotation, scaling, noise addition, cropping and random reordering of points [9]. Furthermore, the increasing sophistication of attacks aimed at evading copyright enforcement underscores the need for more advanced, resilient and robust watermarking strategies.

In response to these challenges, we propose a robust deep neural watermarking framework, specifically tailored for 3D point clouds to enable enhanced copyright protection and ownership verification. Our approach integrates a classical spectral decomposition method—Singular Value Decomposition (SVD)—for imperceptible and resilient watermark embedding, with the extraction capabilities of a modern deep neural network, i.e. PointNet++. For the proposed watermarking approach, detailed in the simplified workflow diagram in Figure 1, Singular Value Decomposition (SVD) is ideal for watermark embedding in 3D point clouds. This is due to its capacity to compactly represent the essential geometric structure of data blocks while preserving perceptual quality. Embedding watermarks in the singular values ensures that the alterations remain imperceptible yet resilient, as these values are less sensitive to minor distortions and noise [10]. This makes SVD a robust choice for spectral-domain watermarking and an effective foundation for further enhancement with deep learning-based extraction.

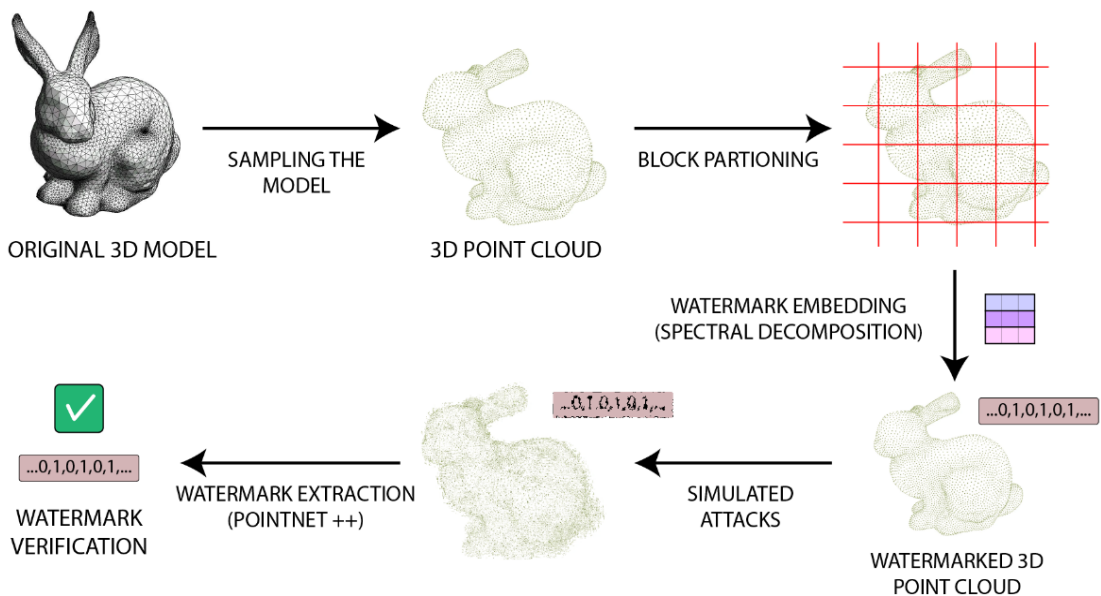

**Figure 1.** Simplified workflow diagram for the deep neural 3D point cloud watermarking

For watermark extraction, we opted for PointNet++, given its widespread adoption as a deep neural network architecture tailored for 3D point clouds. Unlike conventional convolutional neural networks, which rely on structured grid data (such as images), PointNet++ can directly process unordered and irregular point sets [11]. It extends the original PointNet architecture by incorporating a hierarchical feature learning strategy, enabling the extraction of both local and global geometric features from point clouds [12]. This hybrid solution is designed to withstand a comprehensive range of geometric and signal-level attacks, ensuring reliable watermark extraction and ownership verification even under severe adversarial conditions.

The primary aim of this paper is to advance robust copyright protection and ownership verification for 3D point cloud data by introducing a novel hybrid watermarking framework. In contrast to previous approaches that exclusively rely on any one of classical signal processing, deep learning techniques or mesh-based watermarking, our method uniquely integrates block-wise Singular Value Decomposition (SVD) watermark embedding with a deep learning-based PointNet++ decoder, operating directly on 3D point clouds. This approach leverages the complementary strengths of both paradigms and capitalises on the intrinsic properties of point cloud data. The novelty of our method lies in its ability to achieve high-fidelity, imperceptible watermark embedding whilst substantially enhancing resilience to a wide range of geometric and non-geometric attacks, particularly those that disrupt point ordering or cause severe data loss. Through rigorous experimentation on the ModelNet40 dataset[1] that was developed by Zhirong Wu *et al.* [13], it is evident that our hybrid system preserves the integrity and perceptual quality of 3D point clouds whilst enabling highly accurate watermark recovery and ownership verification even under adverse conditions. The purpose of this research is to bridge the gap between classical robustness and modern machine learning adaptability, thereby establishing a new benchmark for practical, scalable and secure 3D content protection in real-world applications.

## 2. Literature Review

The challenge of copyright protection for 3D content has gained increasing research interest in recent years, driven by the rapid growth of applications of 3D models across diverse domains such as entertainment, manufacturing and virtual reality. Digital watermarking remains a cornerstone technology for ensuring copyright protection and content authentication in digital media. Recent advances illustrate how these techniques are evolving to safeguard cutting-edge 3D representations without compromising visual quality, as exemplified by the universal watermarking framework proposed for 3D Gaussian Splatting [14]. While watermarking strategies for 2D images and video are well-matured, their adaptation to three-dimensional (3D) data, particularly point clouds, poses unique technical challenges [15]. Early efforts focused predominantly on polygonal mesh representations, exploiting the connectivity and surface structure of 3D models to embed watermarks [16]. Typical approaches include vertex coordinate modulation, edge flipping and quantisation of surface features, which can be effective for mesh data but are generally inapplicable to point clouds due to their lack of topological information and unordered nature [17]. Before deep learning and machine learning models became popular, 3D watermarking focused on mesh-based and volumetric representations, using spatial [18], frequency [19], or spectral domains [20] for watermark embedding. Several studies have applied vertex perturbation [21], surface normal modification [22] and quantisation techniques [23] to hide information in 3D meshes. These methods, while effective against some basic attacks, often suffer from poor robustness when exposed to aggressive geometric operations, such as resampling, cropping, affine transformations and random point removal. Moreover, many classical algorithms are tightly coupled to mesh connectivity, limiting their applicability to raw point cloud data, which lacks explicit topology [24]. These traditional digital watermarking techniques have demonstrated effectiveness for 2D images and video, have been extensively explored and adapted for 3D data with varying degrees of success, but they are now being replaced by various neural watermarking techniques due to their robustness and resiliency.

### 2.1. Spectral and SVD-Based Watermarking

Spectral decomposition of mesh geometry is a well-established technique, first introduced by Taubin [25] for geometry processing, and still remains highly relevant. This approach has been widely extended for various tasks, such as compression and watermarking of 3D triangle mesh geometry [26]. Spectral methods leverage the mathematical properties of 3D data by transforming models into the frequency or spectral domain, providing more imperceptible and potentially robust spaces for watermark embedding.

Singular Value Decomposition (SVD) is particularly attractive for watermarking as it decomposes matrices constructed from local point cloud coordinates or mesh patches into orthogonal bases and singular values. The singular values encapsulate intrinsic geometric features and small modifications to these values can embed watermark signals with minimal perceptual distortion [27]. SVD-based watermarking schemes are renowned for their stability and resilience to random noise and small-scale distortions due to the global nature of singular values. However, these approaches are vulnerable to geometric transformations—such as global rotation, cropping or reordering of points—that can unpredictably alter singular value distributions and reduce watermark extraction reliability [28]. Some studies attempt to address this limitation by adopting block-wise or local SVD embedding, but robustness remains challenged by non-uniform attacks and point order permutations.

---

[1] Princeton, ModelNet40, Available: https://www.kaggle.com/datasets/balraj98/modelnet40-princeton-3d-object-dataset.

**2.2. Point Cloud-Specific Watermarking**

Unlike meshes, point clouds lack explicit connectivity, making geometric feature extraction and manipulation non-trivial. Early watermarking schemes for point clouds often attempted to extend mesh-based strategies, e.g., directly perturbing coordinates or exploiting spatial distribution statistics [29]. More advanced methods have either incorporated invariant descriptors (e.g., shape distributions, radial-basis features) or utilised graph-based Laplacian spectral transforms to improve attack resilience [30]. Nonetheless, the unordered and irregular sampling of the points implies that even simple operations—such as shuffling, non-uniform sampling, or block cropping—can dramatically degrade classical watermark signals. Robust point cloud watermarking therefore requires embedding strategies that are resilient to such deformations.

**2.3. Deep Learning for 3D Watermark Extraction**

Recent advances in 3D deep learning have provided powerful tools for feature learning, directly from raw point clouds. The introduction of PointNet [31] and its hierarchical extension, PointNet++ [32], allowed networks to consume unordered point sets and aggregate both global and local geometric information through permutation-invariant architectures. Deep learning-based watermark extraction leverages this ability, training neural networks to recover embedded information even after complex attacks. State-of-the-art works have demonstrated that neural networks, when trained with extensive attack augmentation (including noise, rotation, cropping and combined attacks), can generalise well and extract robust features for watermark recovery [33]. Some approaches have combined deep networks with classical watermarking, using neural networks either to guide the embedding or to robustly extract from traditionally embedded watermarks [34].

However, many deep watermarking approaches either focus on mesh data or overlook the spectral and structural properties of point clouds during embedding. Their robustness is often constrained by the limited diversity of attack models used during training, as well as by their reduced ability to generalise to real-world distortions.

**2.4. Robustness and Attack Models**

Robustness evaluation is fundamental to watermarking research, as practical deployment must withstand both intentional removal and accidental degradation. In this context, a wide variety of attack models are used to systematically test the limits of watermark survivability [35]. These include:
   a) Additive noise: Random Gaussian or uniform noise disrupts embedded signals, challenging the noise immunity of the method.
   b) Geometric transformations: Rotations (fixed and arbitrary axes), scaling and translation simulate common manipulations in processing pipelines.
   c) Cropping and point removal: These simulate partial visibility or occlusion, a severe challenge for point cloud watermarking.
   d) Quantisation and signal distortions: Reflect compression, rounding and practical signal-chain degradations.
   e) Shuffling and permutation: Unique to point clouds, this tests the invariance of embedding and extraction to reordering.
   f) Compound attacks: Realistic settings where multiple attacks occur simultaneously.

Most classical approaches, including those based on SVD, DWT or direct spatial embedding, tend to show rapid degradation in watermark recovery rates under adversarial conditions [36]. In contrast, recent studies demonstrate that deep learning-based extraction methods, particularly when trained with attack augmentation and extensive data diversity, can achieve significantly higher watermark recovery rates and improved generalisation to unseen attack scenarios [37]. This robustness is attributed to the ability of neural architecture to learn invariances and subtle correlations in point cloud data that are inaccessible to purely analytical methods.

### 2.5. Related Work

Hybrid watermarking approaches that integrate classical signal processing methods, such as Singular Value Decomposition (SVD), with deep learning-based extraction have emerged to address the limitations of each paradigm when used in isolation. Classical SVD-based methods are well-regarded for their mathematical interpretability, computational efficiency and ability to embed information in a robust yet imperceptible manner by perturbing dominant singular values [38]. However, such schemes can be vulnerable to severe geometric distortions, loss of point correspondence, or non-rigid deformations, all of which are common in 3D data transmission and editing [39].

To overcome these challenges, recent works have proposed hybrid pipelines in which SVD is employed for watermark embedding while a deep neural network that is typically designed for unordered data, is trained to decode the embedded watermark from potentially attacked or altered point clouds [40-41]. This approach leverages the strengths of both domains: SVD's stability and minimal perceptual distortion in the embedding stage and the neural network's capacity for learning invariances to geometric transformations, noise, shuffling and even partial data loss during extraction.

For example, Yang *et al.* [42] propose a system for 3D point cloud copy detection, which first aligns two-point clouds and then computes their similarity using multiple measures. Their method is designed to be robust against manipulations, such as similarity transformations and smoothing. Comprehensive experiments demonstrate its effectiveness under various attack scenarios. Other studies have adapted similar frameworks to mesh and volumetric data, showing that learning-based decoders generalise effectively to a wider range of real-world perturbations when combined with well-understood embedding schemes [43]. This synergy enables the system to maintain watermark integrity in scenarios where either SVD or deep learning alone would fail.

Despite the promise of such hybrid approaches, further research is warranted in several areas: improving generalisation to extremely severe or compound attacks, developing principled adversarial training schemes and establishing rigorous benchmarks on large, diverse 3D datasets.

### 2.6. Research Gap

Clear and persistent research gaps exist as classical SVD-based methods provide strong imperceptibility and moderate robustness but cannot cope with severe or non-uniform geometric attacks. Conversely, purely deep learning-based extraction methods often do not leverage the spectral or geometric structure of point clouds during embedding, and their success heavily depends on the diversity of training data and augmentation schemes [44]. There is a clear need for hybrid frameworks that exploit both spectral-domain embedding and deep learning-based extraction to handle unordered 3D point clouds. This paper addresses this research gap by proposing a novel method that embeds watermarks in the SVD domain of local point cloud blocks and trains PointNet++ to recover watermarks after a diverse set of attacks, achieving superior robustness and fidelity.

### 3. Methodology

This section details our hybrid watermarking framework for 3D point cloud copyright protection and ownership verification. All experiments were conducted on the publicly available Princeton ModelNet40 dataset and early testing was done on ModelNet10 dataset and the popular Stanford Bunny[2] model, comprised of mesh models in '.off' and '.ply' formats. Each mesh was pre-processed by uniform point sampling to obtain point clouds of precisely 1,024 points per sample. For data consistency, each point cloud was centred to zero mean and scaled to a unit maximal Euclidean norm. Any mesh failing to meet these criteria (e.g., due to file corruption or insufficient points) was automatically excluded from the dataset.

---

[2] Greg Turk and Marc Levoy, "Stanford Bunny", Stanford 3D Scanning Repository, 1994, Available: https://graphics.stanford.edu/data/3Dscanrep.

The full dataset was split into distinct training and testing sets, guaranteeing zero overlap to prevent data leakage, as confirmed programmatically. A total of 3,991 training samples (~81.5%) and 908 testing samples (18.5%) were used from the ModelNet40 dataset. The list of valid file paths was dynamically generated, and all point clouds were preloaded into system memory to optimise input/output (I/O) performance during model training and evaluation. All the codes were implemented in Python, leveraging PyTorch and Open3D, as detailed in our implementation[3].

### 3.1. Watermark Embedding and Extraction via Block-SVD

To achieve robust and blind watermarking, we designed a block-based Singular Value Decomposition (block-SVD) mechanism for both embedding and extraction. We lexicographically sorted each 3D point cloud by coordinate, then partitioned it into *n* equal-sized blocks (*n* representing the number of watermark bits; in this case, n=2). Within each block, SVD was performed, and the largest singular value was shifted by a scalar alpha (embedding strength, empirically set to alpha=2.0) according to the assigned watermark bit. This perturbation is imperceptible in most geometric measures but robustly retrievable.

During extraction, the block-wise SVD was recomputed for the potentially attacked point cloud. The difference in the dominant singular values, normalised by alpha, was then thresholded to recover the embedded bits. To quantify fidelity and resilience, we computed bitwise extraction accuracy, intersection-over-union (IoU), bit error rate (BER), peak signal-to-noise ratio (PSNR) and symmetric Chamfer distance.

### 3.2. Data Augmentation

To improve the robustness and generalisability of the deep learning-based decoder, we applied a series of data augmentation techniques to the point clouds during training. Each time a point cloud is sampled from the training set, it is subjected to a stochastic augmentation pipeline. The augmentations applied (in random order and combination) include Gaussian noise ($\sigma$=0.01), random scaling (0.95-1.05), random rotation, random dropout (10%) and re-normalisation. These augmentations are only applied during training and not during validation or testing. This on-the-fly data augmentation strategy increases the diversity of the training data and helps prevent overfitting, enabling the PointNet++ decoder to learn features that are invariant to common geometric transformations, noise and sampling variations.

### 3.3. Deep Learning-Based Watermark Decoding (PointNet++)

In addition to the classical SVD-based decoder which we used as the baseline, we trained a deep neural network for robust watermark extraction. The network is based on the PointNet++ architecture, which consists of a hierarchy of set-abstraction modules with local and global feature aggregation [46]. The encoder receives a normalised point cloud and predicts the watermark bits as output logits. The model was trained using binary cross-entropy loss, with Adam optimisation and a learning rate scheduler to facilitate convergence. Training was performed for up to 150 epochs, with a batch size of 32 and an initial learning rate of 2e-3, using mini-batch processing and validation after each epoch. The bitwise accuracy and per-bit performance were tracked throughout training. The best-performing PointNet++ checkpoint (the model instance that achieved the highest validation accuracy, saved as *best_model.pt*) was selected for the final evaluation.

### 3.4. Adversarial Attacks

To systematically evaluate the watermark robustness, we implemented a comprehensive suite of both geometric and non-geometric attacks. These attacks were carefully selected to be representative of real-world 3D processing pipelines, encompassing a wide range of transformations and degradations that 3D point clouds commonly undergo in practical applications. The goal was to rigorously test the watermark's ability to withstand diverse forms of manipulation and distortion, ensuring its persistence and detectability even under challenging conditions. Table 1 details the specific types of attacks employed and their associated parameter values, providing a clear overview of the test conditions.

---

[3] Khandoker Ashik Uz Zaman, Mahdi H. Miraz, Mohammad Zahangir Alam and Mohammad Ali, "Deep Neural Watermarking", Zenodo, 2025, DOI: 10.5281/zenodo.15756083, Available: https://zenodo.org/records/15756083.

Table 1. Attacks used for model evaluation and their parameters

| Type of Attack | Parameters |
| --- | --- |
| Additive Gaussian Noise | σ=0.01 |
| Additive Gaussian Noise | σ=0.03 |
| Gaussian Smoothing | k=16, σ=0.05 |
| Random isotropic scaling | Range: 0.8, 1.2 |
| Random Rotation | Random |
| Random rotation (about fixed and arbitrary axes) | Random |
| Random translation | Max shift: 0.05 |
| Point dropout | Removal of up to 10% points |
| Shuffle | Random |
| Random cropping | 70% retention |
| Affine distortion | Axis-wise scaling |
| Quantisation | Step size: 0.01 |
| Jittering | Gaussian, σ=0.005 |
| Chunk removal, Gaussian smoothing | Fraction: 20% |
| Combined attacks (Noise & Dropout) | σ=0.02, 0.15 |

All attacks were applied to watermarked test point clouds prior to watermark extraction. Both the SVD and PointNet++ models were utilised to extract the watermark from which we calculated the bitwise accuracy, intersection-over-union (IoU) between ground truth and predicted watermark, bit error rate (BER), Chamfer distance and peak signal-to-noise ratio (PSNR), to assess geometric distortion. Aggregate statistics, including mean and standard deviation for each metric under every attack scenario, were compiled, refer to section 4 (Results and Discussion). Detailed accuracy tables and bar plots were generated to visualise the results, and ownership verification was tested via receiver operating characteristic (ROC) analysis to demonstrate the discriminative capability of the model in authenticating watermarked point clouds. All the codes were implemented in Python using PyTorch, NumPy and Open3D. All the experiments were run on google colab pro, using an NVIDIA L4 GPU with high-RAM acceleration.

## 4. Results and Discussion

Prior to training, dataset integrity and class balance were confirmed. The dataset comprised 4,899 unique point clouds, partitioned into 3,991 for training and 908 for testing, with each normalised to zero mean and unit scale (see Table 1). Eight unique watermark bit patterns were present in the training set, ensuring sufficient diversity for effective learning.

The PointNet++ decoder was trained for 150 epochs using the Adam optimiser and a ReduceLROnPlateau learning-rate schedule, which led to a steady reduction in loss. The final train loss reached as low as 0.08 and validation loss as low as 0.07. Figure 2A presents the evolution of training and validation loss across epochs, indicating rapid convergence and generalisation after epoch 50. Figure 2B depicts the corresponding training and validation accuracy, which plateaued at approximately 97% by epoch 130, confirming stable model learning.

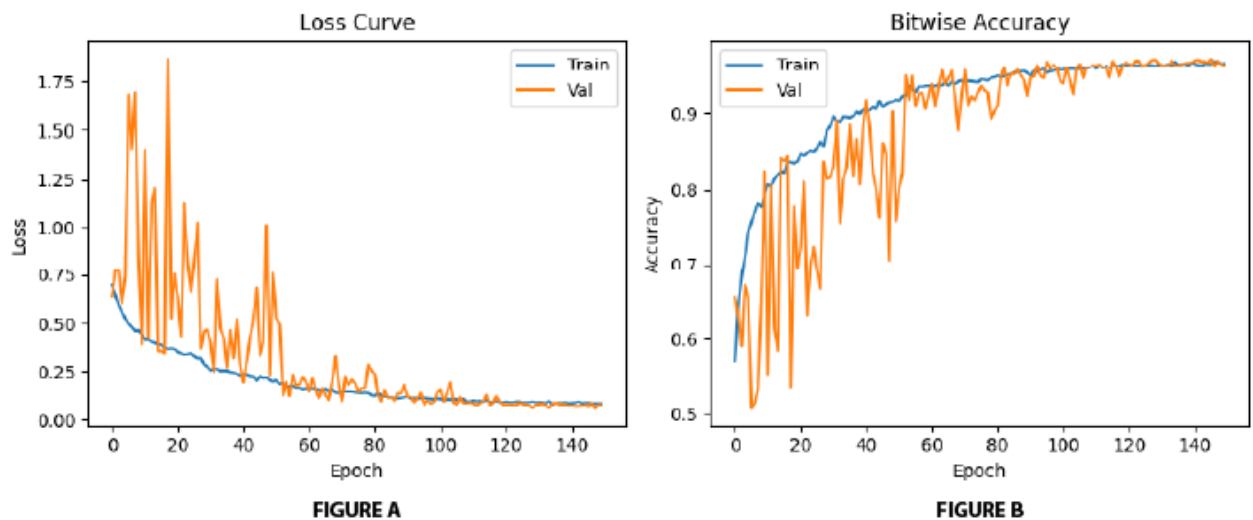

**Figure 2.** A) Train and validation loss curve, B) Train and validation bitwise accuracy

Per-bit validation accuracy for all three watermark bits (see Figure 3) was consistently high, with each bit exceeding 96% accuracy by the end of training. For randomly selected test samples, the model achieved perfect bitwise decoding in several cases, demonstrating high fidelity in watermark embedding and decoding on clean data.

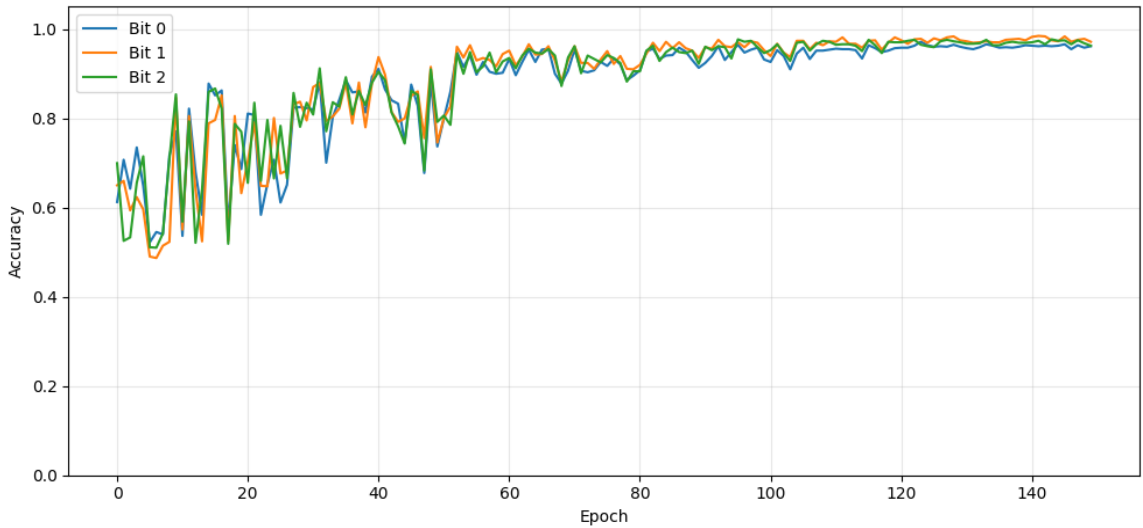

**Figure 3.** Per-bit validation accuracy curve

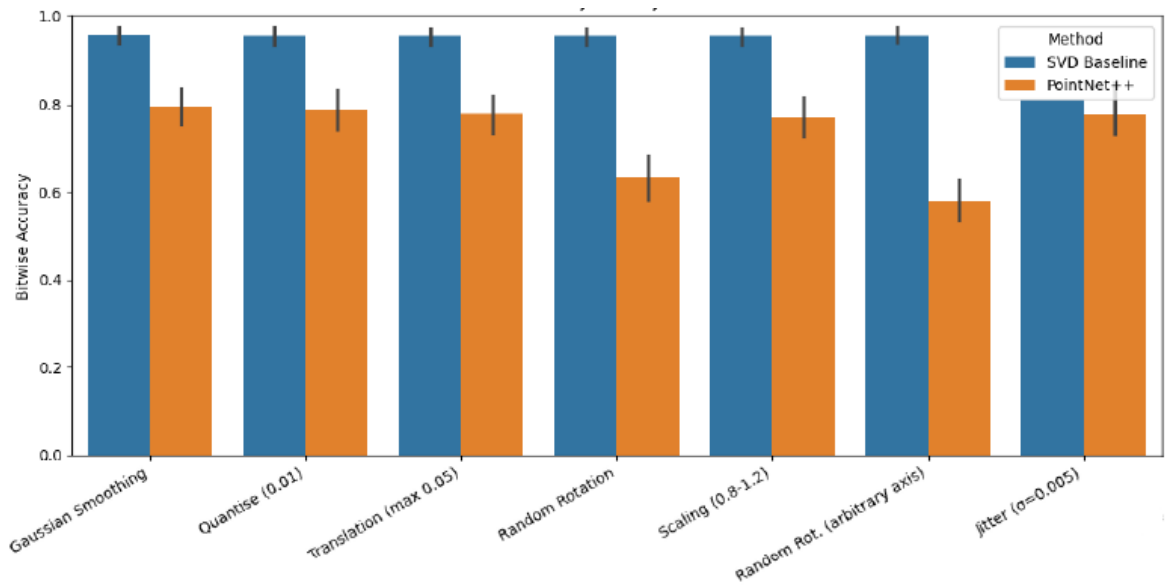

**Figure 4.** Watermark recovery accuracy under 3D attacks Part A

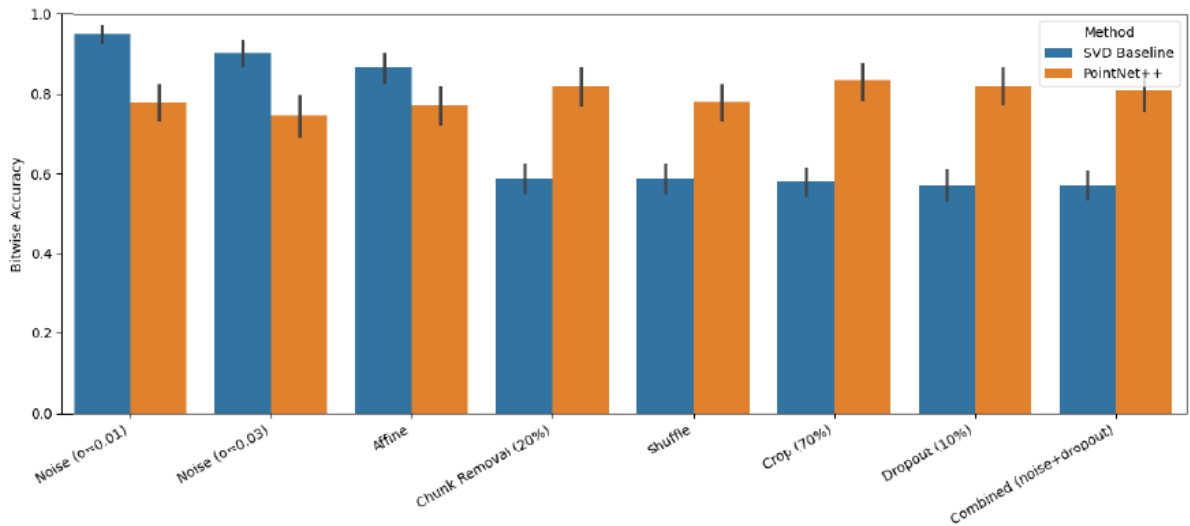

**Figure 5.** Watermark recovery accuracy under 3D attacks Part B

To comprehensively evaluate watermark recovery, we subjected the watermarked point clouds to a wide spectrum of geometric and non-geometric attacks. Figures 4 and 5 illustrate the recovery accuracies for both the block-wise SVD baseline and the proposed deep learning model under diverse attack scenarios. Table 2 further quantifies the accuracy gap between these two approaches across all attack types.

Table 2. Accuracy gaps between baseline and proposed model under different attack scenarios

| Type of Attack | SVD | PointNet++ | Accuracy Gap |
|---|---|---|---|
| Additive Gaussian Noise (0.01) | 0.950 | 0.778 | -0.175 |
| Additive Gaussian Noise (0.03) | 0.903 | 0.777 | -0.173 |
| Gaussian Smoothing | 0.957 | 0.795 | -0.162 |
| Random isotropic scaling | 0.955 | 0.772 | -0.183 |
| Random Rotation | 0.955 | 0.633 | -0.322 |
| Random rotation (arbitrary axes) | 0.955 | 0.582 | -0.373 |
| Random translation | 0.955 | 0.780 | -0.175 |
| Quantisation | 0.955 | 0.788 | -0.167 |
| Jittering | 0.953 | 0.778 | -0.175 |
| Affine distortion | 0.867 | 0.773 | -0.093 |
| Shuffle | 0.588 | 0.780 | +0.192 |
| Random Cropping | 0.580 | 0.818 | +0.230 |
| Point dropout | 0.572 | 0.820 | +0.248 |
| Chunk removal | 0.588 | 0.818 | +0.230 |
| Combined attacks (Noise & Dropout) | 0.572 | 0.810 | +0.238 |

Beyond simple accuracy, we assessed robustness using Chamfer Distance (CD), Bit Error Rate (BER) and Peak Signal-to-Noise Ratio (PSNR)—standard metrics for 3D fidelity and watermark reliability. Mild attacks, such as Gaussian smoothing, quantisation and translation, resulted in low Chamfer Distances (~0.10) and high PSNR values (≥21 dB), indicating minimal geometric distortion and strong watermark integrity. Correspondingly, both the SVD method and deep model maintained bit error rates (BERs) below 5% in these scenarios. However, under more destructive attacks, such as random shuffling, cropping and point dropout, the block-wise SVD method exhibited sharp increases in BER (up to 42% in some cases) and significant drops in PSNR, reflecting the loss of watermark information. In these challenging cases, the deep learning decoder consistently delivered lower BER and higher PSNR than the SVD baseline, evidencing its greater resilience. For instance, with a 10% point dropout attack, the deep model achieved a BER of 18% (accuracy 82%), compared to 43% BER (accuracy 57%) for SVD. Mean BER across all attacks was 18% for the deep model and 22% for SVD, while mean PSNR values were generally higher for the neural approach under severe distortions. Across all attack types, the mean SVD accuracy was approximately 82%, while the deep-learning model averaged 77%. However, the deep model consistently outperformed SVD under severe distortions, where robustness is most critical. Despite both methods being challenged by compounded attacks (such as combined noise and dropout, or random cropping), the deep model reliably narrowed the performance gap. These trends (detailed in Table 3) are also reflected in the average Chamfer Distance, which remained acceptably low for both models under mild attacks but notably increased under severe geometric manipulations.

Table 3. Chamfer distance, PSNR and BER comparison under different attack scenarios

| Type of Attack | Chamfer_SVD | Chamfer_DL | PSNR_SVD | PSNR_DL | BER_SVD | BER_DL |
|---|---|---|---|---|---|---|
| Additive Gaussian Noise (0.01) | 0.105 | 0.105 | 21.9 | 21.9 | 0.05 | 0.22 |
| Additive Gaussian Noise (0.03) | 0.109 | 0.109 | 20.8 | 20.7 | 0.10 | 0.26 |
| Gaussian Smoothing | 0.104 | 0.104 | 22.1 | 22.1 | 0.04 | 0.21 |
| Random isotropic scaling | 0.105 | 0.105 | 22.0 | 22.0 | 0.05 | 0.23 |
| Random Rotation | 0.266 | 0.255 | 7.89 | 8.34 | 0.05 | 0.22 |
| Random rotation (arbitrary axes) | 0.328 | 0.339 | 7.71 | 7.35 | 0.05 | 0.42 |
| Random translation | 0.105 | 0.105 | 22.0 | 22.0 | 0.05 | 0.37 |
| Quantisation | 0.105 | 0.105 | 22.0 | 22.0 | 0.05 | 0.21 |
| Jittering | 0.105 | 0.105 | 22.0 | 22.0 | 0.47 | 0.22 |
| Affine distortion | 0.111 | 0.110 | 21.7 | 21.5 | 0.13 | 0.23 |
| Shuffle | 0.105 | 0.105 | 4.14 | 4.15 | 0.41 | 0.22 |
| Random Cropping | 0.110 | 0.110 | 4.12 | 4.13 | 0.42 | 0.17 |
| Point dropout | 0.106 | 0.106 | 4.15 | 4.14 | 0.43 | 0.18 |
| Chunk removal | 0.108 | 0.107 | 4.14 | 4.13 | 0.41 | 0.18 |
| Combined attacks | 0.109 | 0.109 | 4.19 | 4.17 | 0.43 | 0.19 |

The summary tables and bar plot indicate that the SVD method demonstrates high robustness to mild attacks such as Gaussian smoothing, quantisation and translations, achieving approximately 95% accuracy in these scenarios. Although the deep learning-based approach, i.e. PointNet++, typically matches or slightly lags behind the SVD baseline under mild distortions, it significantly outperforms SVD when subjected to severe attacks that disrupt the spatial structure of the point cloud, including point dropout, random shuffling or cropping. For example, in the case of a 10% dropout attack, the deep learning model attained approximately 82% accuracy, compared with only 57% for SVD. Across all attack types, the mean SVD accuracy was around 82%, while the deep learning model averaged about 77%. Despite both approaches finding combined and random shuffle/crop attacks particularly challenging, the deep model consistently narrows the performance gap in these difficult conditions.

For ownership verification, the discriminative capacity of the deep decoder was assessed via ROC analysis (see Figure 6). The area under the ROC curve (AUC) reached 0.67, demonstrating that the model can distinguish between authentic and random watermarks with reasonable reliability; however further improvement is needed.

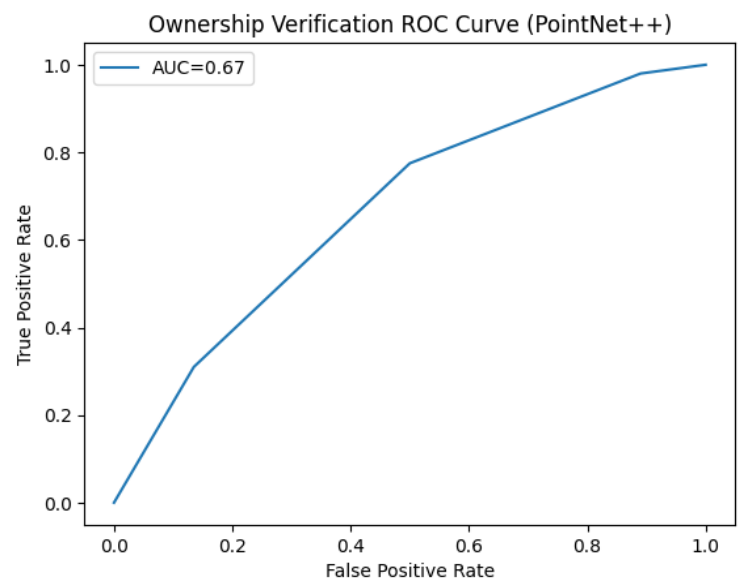

**Figure 6.** Ownership Verification ROC Curve (PointNet++)

The results demonstrate that block-wise SVD watermarking is robust to a wide range of mild geometric distortions, but sensitive to attacks that break point cloud ordering or remove significant portions of the data. The deep learning-based decoder shows excellent generalisation, with high per-bit accuracy, and is particularly effective at recovering watermarks from attacked clouds that defeat SVD-based extraction. Nonetheless, the remaining challenges include enhancing resilience to the most destructive attacks, e.g. chunk removal, random crop, etc. and further improving the ownership verification pipeline, where AUC can likely be boosted via ensemble methods or larger and more diverse training data. To address this, one approach could be to use both SVD and DL models and compare their predictions to determine the final ownership verification, while another approach could be to use a better deep learning model. However, the latter would require substantially more training data, computational power and time.

## 5. Concluding Discussion and Future Research Directions

In this work, we introduced a robust watermarking framework for 3D point clouds, combining classical block-wise SVD-based embedding with a deep learning-based decoder. Through extensive experiments and a diverse suite of geometric and non-geometric attacks, we demonstrated that the proposed method achieves high watermark-recovery accuracy and strong generalisation, particularly when using the PointNet++ architecture for watermark extraction. The SVD baseline offers remarkable robustness to mild perturbations, while the deep network excels under severe or disordered conditions that typically defeat conventional approaches. The experimental results validate the effectiveness of our joint methodology and highlight its potential for practical, secure ownership verification in 3D data.

While our approach demonstrates promising results, several challenges remain. First, enhancing resilience to highly destructive attacks—such as random cropping, chunk removal and point-cloud shuffling—requires further research, potentially by developing advanced architectures tailored for unordered or incomplete point sets. Future work could explore more robust hybrid verification pipelines that combine the strengths of both SVD and deep learning models, for example by fusing or assembling their outputs to improve reliability. Additionally, scaling up to a larger and more varied datasets, incorporating self-supervised or contrastive learning strategies and optimising computational efficiency for deployment are valuable avenues. Ultimately, strengthening the ownership-verification pipeline and further boosting the AUC, perhaps through model ensembles or uncertainty quantification, remain key directions for advancing practical and robust 3D watermarking solutions.

**CRediT Author Contribution Statement**

Khandoker Ashik Uz Zaman: Methodology, Software, Formal analysis, Investigation, Data curation, Writing – original draft, Visualisation; Mohammad Zahangir Alam: Writing – original draft, Writing – review & editing, Formal analysis; Mohammad N.M. Ali: Supervision, Writing – review & editing; Mahdi H. Miraz: Conceptualisation, Supervision, Project administration, Funding acquisition, Writing – review & editing.

**Acknowledgement**

This research is financially supported by Xiamen University Malaysia (Project codes: XMUMRF/2021-C8/IECE/0025 and XMUMRF/2022-C10/IECE/0043).

**References**

[1] Shen Ying, Peter van Oosterom and Hongchao Fan, "New Techniques and Methods for Modelling, Visualization, and Analysis of a 3D City", *Journal of Geovisualization and Spatial Analysis (JGSA)*, Print ISSN: 2509-8810, Online ISSN: 2509-8829, Vol. 7, No. 2, December 2023, Art. No. 26, Published by Springer Nature, DOI: 10.1007/s41651-023-00157-x, Available: https://link.springer.com/article/10.1007/s41651-023-00157-x.

[2] Zahra Rezaei, Mohammad Hossein Vahidnia, Hossein Aghamohammadi, Zahra Azizi and Saeed Behzadi, "Digital twins and 3D information modeling in a smart city for traffic controlling: A review", *Journal of Geography and Cartography*, Online ISSN: 2578-1979, Vol. 6, No. 1, 4 January 2023, Art. No. 1865, Published by EnPress Publisher, DOI: 10.24294/jgc.v6i1.1865, Available: https://systems.enpress-publisher.com/index.php/JGC/article/view/1865.

[3] Ruijun Liu, Haisheng Li and Zhihan Lv, "Modeling Methods of 3D Model in Digital Twins", *Computer Modeling in Engineering and Sciences,* Print ISSN: 1526-1492, Online ISSN: 1526-1506, Vol. 136, No. 2, 6 February 2023, pp. 985–1022, Published by Tech Science Press, DOI: 10.32604/cmes.2023.023154, Available: https://www.techscience.com/CMES/v136n2/51588.

[4] Ziyi Zhang, Liming Zhang, Pengbin Wang, Mingwang Zhang and Tao Tan, "Robust watermarking algorithm based on Mahalanobis distance and ISS feature point for 3D point cloud data", *Earth Science Informatics*, Print ISSN: 1865-0473, Online ISSN: 1865-0481, Vol. 17, No. 1, February 2024, pp. 783–796, Published by Springer Nature, DOI: 10.1007/s12145-023-01206-1, Available: https://link.springer.com/article/10.1007/s12145-023-01206-1.

[5] Saifullahi Aminu Bello, Shangshu Yu, Cheng Wang, Jibril Muhammad Adam and Jonathan Li, "Review: Deep Learning on 3D Point Clouds", *Remote Sensing*, Online ISSN: 2072-4292, Vol. 12, No. 11, 28 May 2020, Art. No. 1729, Published by MDPI, DOI: 10.3390/rs12111729 , Available: https://www.mdpi.com/2072-4292/12/11/1729.

[6] Bisheng Yang, Niels Haala and Zhen Dong, "Progress and perspectives of point cloud intelligence", *Geo-spatial Information Science*, Print ISSN: 1009-5020, Online ISSN: 1993-5153, Vol. 26, No. 2, April 2023, pp. 189–205, Published by Taylor & Francis Online, DOI: 10.1080/10095020.2023.2175478, Available: https://www.tandfonline.com/doi/full/10.1080/10095020.2023.2175478.

[7] Felipe A.B.S. Ferreira and Juliano B. Lima, "A robust 3D point cloud watermarking method based on the graph Fourier transform", *Multimedia Tools and Applications*, Online ISSN: 1573-7721, Vol. 79, No. 3, January 2020, pp. 1921–1950, Published by Springer Nature, DOI: 10.1007/s11042-019-08296-4, Available: https://link.springer.com/article/10.1007/s11042-019-08296-4.

[8] Abderrazzaq Kharroubi, Florent Poux, Zouhair Ballouch, Rafika Hajji and Roland Billen, "Three Dimensional Change Detection Using Point Clouds: A Review", *Geomatics*, Online ISSN: 2673-7418, Vol. 2, No. 4, 17 October 2022, pp. 457–485, Published by MDPI, DOI: 10.3390/geomatics2040025, Available: https://www.mdpi.com/2673-7418/2/4/25.

[9] Cheng Wei, Yang Wang, Kuofeng Gao, Shuo Shao, Yiming Li *et al.*, "PointNCBW: Towards Dataset Ownership Verification for Point Clouds via Negative Clean-label Backdoor Watermark", *IEEE Transactions on Information Forensics and Security*, Print ISSN: 1556-6013, Online ISSN: 1556-6021, Vol. 20, 6 November 2024, pp. 191–206, Published by IEEE, DOI: 10.1109/TIFS.2024.3492792, Available: https://ieeexplore.ieee.org/document/10745757.

[10] Sunil Gupta, Kamal Saluja, Vikas Solanki, Kushwant Kaur, Parveen Singla *et al.*, "Efficient methods for digital image watermarking and information embedding", *Measurement: Sensors*, Online ISSN: 2665-9174, Vol. 24, December 2022, Art. No. 100520, Published by Elsevier BV, DOI: 10.1016/j.measen.2022.100520, Available: https://www.sciencedirect.com/science/article/pii/S2665917422001544.

[11] Yang Chen, Guanlan Liu, Yaming Xu, Pai Pan and Yin Xing, "PointNet++ Network Architecture with Individual Point Level and Global Features on Centroid for ALS Point Cloud Classification", *Remote Sensing*, Online ISSN: 2072-4292, Vol. 13, No. 3, 29 January 2021, Art. No. 472, Published by MDPI, DOI: 10.3390/rs13030472, Available: https://www.mdpi.com/2072-4292/13/3/472.

[12] Yanchao Lian, Tuo Feng and Jinliu Zhou, "A Dense Pointnet++ Architecture for 3D Point Cloud Semantic Segmentation", in *Proceedings of the 2019 IEEE International Geoscience and Remote Sensing Symposium (IGARSS 2019)*, 28 July–2 August 2019, Yokohama, Japan, Print ISSN: 2153-6996, Online ISSN: 2153-7003, E-ISBN: 978-1-5386-9154-0, Print ISBN: 978-1-5386-9155-7, pp. 5061–5064, Published by IEEE, DOI: 10.1109/IGARSS.2019.8898177, Available: https://ieeexplore.ieee.org/document/8898177.

[13] Zhirong Wu, Shuran Song, Aditya Khosla, Fisher Yu, Linguang Zhang *et al.*, "3D ShapeNets: A Deep Representation for Volumetric Shapes", in *Proceedings of the 2015 IEEE Conference on Computer Vision and Pattern Recognition (CVPR)*, 7–12 June 2015, Boston, MA, USA, pp. 1912–1920, Published by IEEE Computer Society, DOI: 10.1109/CVPR.2015.7298801, Available: https://ieeexplore.ieee.org/document/7298801.

[14] Lijiang Li, Jinglu Wang, Xiang Ming and Yan Lu, "GS-Marker: Generalizable and Robust Watermarking for 3D Gaussian Splatting", *arXiv preprint*, 24 March 2025, DOI: 10.48550/arXiv.2503.18718, Available: https://arxiv.org/abs/2503.18718.

[15] Cong Wang, Yang Luo, Ke Wang, Yanfei Cao, Xiangzhi Tao *et al.*, "Breaking barriers in 3D point cloud data processing: A unified system for efficient storage and high-throughput loading", *Expert Systems with Applications: An International Journal*, Print ISSN: 0957-4174, Online ISSN: 1873-6793, Vol. 277, No. C, 5 June 2025, Art. No. 126983, Published by Elsevier Ltd., DOI: 10.1016/j.eswa.2025.126983, Available: https://www.sciencedirect.com/science/article/abs/pii/S0957417425006050.

[16] Thomas Harte and Adrian G. Bors, "Watermarking 3D models", in *Proceedings of the 2002 IEEE International Conference on Image Processing (ICIP 2002)*, 22–25 September 2002, Rochester, NY, USA, Vol. 3, Print ISSN: 1522-4880, Online ISSN: 2381-8549, pp. 661–664, Published by IEEE, DOI: 10.1109/ICIP.2002.1039057, Available: https://ieeexplore.ieee.org/document/1039057.

[17] Jian Liu, Yaije Yang, Douli Ma, Yinghui Wang, Zhigeng Pan *et al.*, "A Watermarking Method for 3D Models Based on Feature Vertex Localization", *IEEE Access*, Online ISSN: 2169-3536, Vol. 6, 30 September 2018, pp. 56122–56134, Published by IEEE, DOI: 10.1109/ACCESS.2018.2872783, Available: https://ieeexplore.ieee.org/document/8478155.

[18] Xinyu Wang, Yongzhao Zhan and Shun Du, "A Non-blind Robust Watermarking Scheme for 3D Models in Spatial Domain", in *Electrical Engineering and Control: Selected Papers from the 2011 International Conference on Electric and Electronics (EEIC 2011)*, 20–22 June 2011, Nanchang, China, Vol. 2, Berlin, Heidelberg: Springer Berlin Heidelberg, 2011, Print ISBN: 978-3-642-21764-7, Online ISBN: 978-3-642-21765-4, Series Print ISSN: 1876-1100, Series Online ISSN: 1876-1119, pp. 621–628, DOI: 10.1007/978-3-642-21765-4_76, Available: https://link.springer.com/chapter/10.1007/978-3-642-21765-4_76.

[19] Ryutarou Ohbuchi, Akio Mukaiyama and Shigeo Takahashi, "A Frequency-Domain Approach to Watermarking 3D Shapes", *Computer Graphics Forum*, Online ISSN: 1467-8659, Print ISSN: 0167-7055, Vol. 21, No. 3, pp. 373–382, September 2002, Published by Blackwell Publishing, Inc., Oxford, UK, DOI: 10.1111/1467-8659.t01-1-00597, Available: https://onlinelibrary.wiley.com/doi/10.1111/1467-8659.t01-1-00597.

[20] Ryutarou Ohbuchi, Shigeo Takahashi, Takahiko Miyazawa and Akio Mukaiyama, *"Watermarking 3D polygonal meshes in the mesh spectral domain", in Proceedings of Graphics Interface 2001 (GI '01)*, 7-9 June 2001, Ottawa, Ontario, Canada, ISBN: 978-0-9688808-0-7, pp. 9–17, Published by Canadian Information Processing Society, DOI: 10.20380/GI2001.02, Available: https://dl.acm.org/doi/10.5555/780986.780989.

[21] Huangxinxin Xu, Fazhi He, Linkun Fan and Junwei Bai, "D3AdvM: A direct 3D adversarial sample attack inside mesh data", *Computer Aided Geometric Design*, Print ISSN: 0167-8396, Online ISSN: 1879-2332, Vol. 97, 1 August 2022, Art. No. 102122, Published by Elsevier, DOI: 10.1016/j.cagd.2022.102122, Available: https://linkinghub.elsevier.com/retrieve/pii/S0167839622000589.

[22] Emil Praun, Hugues Hoppe and Adam Finkelstein, "Robust mesh watermarking", in *Proceedings of the 26th Annual Conference on Computer Graphics and Interactive Techniques (SIGGRAPH '99)*, 8-13 August 1999, Los Angeles, California, USA, ISBN: 978-0-201-48560-8, pp. 49–56, Published by ACM Press/Addison-Wesley Publishing Co., DOI: 10.1145/311535.311540, Available: https://dl.acm.org/doi/10.1145/311535.311540.

[23] Franck Cayre and Benoît Macq, "Data hiding on 3-D triangle meshes", *IEEE Transactions on Signal Processing*, Vol. 51, No. 4, 30 April 2003, pp. 939–949, Print ISSN: 1053-587X, Online ISSN: 1941-0476, Published by IEEE, DOI: 10.1109/TSP.2003.809380, Available: https://ieeexplore.ieee.org/document/1188740.

[24] Narendra Modigari, M. L. Valarmathi and L. Jani Anbarasi, "Watermarking techniques for three-dimensional (3D) mesh models: a survey", *Multimedia Systems*, Print ISSN: 0942-4962, Online ISSN: 1432-1882, Vol. 28, No. 2, April 2022, pp. 623–641, Published by Springer Nature, DOI: 10.1007/s00530-021-00860-z, Available: https://link.springer.com/article/10.1007/s00530-021-00860-z.

[25] Gabriel Taubin, Tong Zhang and Gene H. Golub, "Optimal surface smoothing as filter design", in Lecture Notes in *Computer Science: Proceedings of the 4th European Conference on Computer Vision (ECCV '96)*, 14–18 April 1996, Cambridge, UK, Vol. 1064, Springer, Berlin, Heidelberg, 1996, Series Print ISSN: 0302-9743, Series Online ISSN: 1611-3349, Print ISBN: 978-3-540-61122-6, Online ISBN: 978-3-540-49949-7, pp. 283–292, DOI: 10.1007/BFb0015544, Available: https://link.springer.com/chapter/10.1007/bfb0015544.

[26] Franck Cayre, Pedro Rondao-Alface, Frédéric Schmitt, Benoît Macq and Henri Maître, "Application of spectral decomposition to compression and watermarking of 3D triangle mesh geometry", *Signal Processing: Image Communication*, Print ISSN: 0923-5965, Online ISSN: 1879-2677, Vol. 18, No. 4, 1 April 2003, pp. 309–319, Published by Elsevier B.V., DOI: 10.1016/S0923-5965(02)00147-9, Available: https://www.sciencedirect.com/science/article/abs/pii/S0923596502001479.

[27] Zhou Zude., Ai Qingsong. and Liu Quan, "A SVD-based Digital Watermarking Algorithm for 3D Mesh Models", in *Proceedings of the 8th International Conference on Signal Processing (ICSP 2006)*, 16–20 November 2006, Guilin, China, Print ISSN: 2164-5221, Online ISSN: 2164-523X, Print ISBN: 0-7803-9736-3, Published by IEEE, DOI: 10.1109/ICOSP.2006.345963, Available: https://ieeexplore.ieee.org/document/4129655.

[28] Ming Tang and Fuken Zhou, "A robust and secure watermarking algorithm based on DWT and SVD in the fractional order Fourier transform domain", *Array*, Online ISSN: 2590-0056, Vol. 15, 1 September 2022, Art. No. 100230, Published by Elsevier, DOI: 10.1016/j.array.2022.100230, Available: https://www.sciencedirect.com/science/article/pii/S2590005622000698.

[29] Divya and Sreeleja N. Unnithan, "A Review on 3D Image Watermarking and Quality Assessment", *International Advanced Research Journal in Science, Engineering and Technology*, Print ISSN: 2394-1588, Online ISSN: 2393-8021, Vol. 6, No. 3, March 2019, pp. 147–151, Published by IARJSET, DOI: 10.17148/IARJSET.2019.6322, Available: https://iarjset.com/wp-content/uploads/2019/04/IARJSET.2019.6322.pdf.

[30] Ruiguo Yang, Xinhui Han, Wenfa Qi and Wei Hu, "Robust Watermark Imaging via Graph-signal Optimization", in *Proceedings of the 2023 Asia Pacific Signal and Information Processing Association Annual Summit and Conference (APSIPA ASC),* 31 October–3 November 2023, Taipei, Taiwan, Print ISSN: 2640-009X, Online ISSN: 2640-0103, E-ISBN: 979-8-3503-0067-3, Print ISBN: 979-8-3503-0068-0, pp. 1451–1457, Published by IEEE, DOI: 10.1109/APSIPAASC58517.2023.10317390, Available: https://ieeexplore.ieee.org/document/10317390.

[31] Charles Ruizhongtai Qi, Hao Su, Mo Kaichun and Leonidas J. Guibas, "PointNet: Deep Learning on Point Sets for 3D Classification and Segmentation", in *Proceedings of the 2017 IEEE Conference on Computer Vision and Pattern Recognition (CVPR)*, 21–26 July 2017, Honolulu, HI, USA, Print ISSN: 1063-6919, Print ISBN: 978-1-5386-0458-8, E-ISBN: 978-1-5386-0457-1, pp. 77–85, DOI: 10.1109/CVPR.2017.16, Available: https://ieeexplore.ieee.org/document/8099499.

[32] Charles Ruizhongtai Qi, Li Yi, Hao Su and Leonidas J. Guibas, "PointNet++: Deep Hierarchical Feature Learning on Point Sets in a Metric Space", in *Advances in Neural Information Processing Systems 30 (NeurIPS 2017)*, 4–9 December 2017, Long Beach, CA, USA, ISBN: 978-1-5108-6096-4, pp. 5105–5114, Published by Curran Associates, Inc., DOI: 10.48550/arXiv.1706.02413, Available: https://dl.acm.org/doi/10.5555/3295222.3295263.

[33] Wentao Qu, Jing Wang, YongShun Gong, Xiaoshui Huang and Liang Xiao, "An End-to-End Robust Point Cloud Semantic Segmentation Network with Single-Step Conditional Diffusion Models", in *Proceedings of the IEEE/CVF Conference on Computer Vision and Pattern Recognition (CVPR '25)*, 11–15 June 2025, Nashville, USA, pp. 27325–27335, DOI: 10.48550/arXiv.2411.16308, Available: https://ui.adsabs.harvard.edu/abs/2024arXiv241116308Q/abstract.

[34] Saoussen Ben Jabra and Mohamed Ben Farah, "Deep Learning-Based Watermarking Techniques Challenges: A Review of Current and Future Trends", *Circuits, Systems, and Signal Processing*, Print ISSN: 0278-081X, Online ISSN: 1531-5878, Vol. 43, No. 7, July 2024, pp. 4339–4368, Published by Springer Nature, DOI: 10.1007/s00034-024-02651-z, Available: https://link.springer.com/article/10.1007/s00034-024-02651-z.

[35] Hung-Hsu Tsai, Yu-Jie Jhuang and Yen-Shou Lai, "An SVD-based image watermarking in wavelet domain using SVR and PSO", *Applied Soft Computing*, Print ISSN: 1568-4946, Online ISSN: 1872-9681, Vol. 12, No. 8, pp. 2442–2453, 1 August 2012, Published by Elsevier BV, DOI: 10.1016/j.asoc.2012.02.021, https://www.sciencedirect.com/science/article/abs/pii/S156849461200110X.

[36] Muhammad Romadhona Kusuma and Supriadi Panggabean, "Robust Digital Image Watermarking Using DWT, Hessenberg, and SVD for Copyright Protection", in *IJACI: International Journal of Advanced Computing and Informatics*, Online ISSN: 3089-7483, Vol. 2, No. 1, 10 August 2025, pp. 41–52, Published by Cendekia Publisher, DOI: 10.71129/ijaci.v2i1.pp41-52, Available: https://journal.cendekiajournal.com/ijaci/article/view/24.

[37] Himanshu Kumar Singh and Amit Kumar Singh, "Digital image watermarking using deep learning", *Multimedia Tools and Applications,* Online ISSN: 1573-7721, Vol. 83, January 2024, pp. 2979–2994, Published by Springer Nature, DOI: 10.1007/s11042-023-15750-x, Available: https://link.springer.com/article/10.1007/s11042-023-15750-x.

[38] Yun Zhu, "Analysis and Simulation on SVD-Based 3D Mesh Digital Watermark Algorithm", *Advanced Materials Research,* Vol. 846–847, November 2013, pp. 1052–1055, Print ISSN: 1022-6680, Online ISSN: 1662-8985, Published by Trans Tech Publications Ltd., DOI: 10.4028/www.scientific.net/AMR.846-847.1052, Available: https://www.scientific.net/AMR.846-847.1052.

[39] Zhongpai Gao, Alex Hwang, Guangtao Zhai and Eli Peli, "Correcting geometric distortions in stereoscopic 3D imaging", *PLOS One*, Vol. 13, No. 10, 8 October 2018, Art. No. e0205032, DOI: 10.1371/journal.pone.0205032, Available: https://journals.plos.org/plosone/article?id=10.1371/journal.pone.0205032.

[40] Swanirbhar Majumder, Tirtha Shankar Das, Vijay H. Mankar and Subir K. Sarkar, "SVD and Neural Network Based Watermarking Scheme", in *Proceedings of the Information Processing and Management – BAIP 2010, Communications in Computer and Information Science,* 26-27 March, 2010, Vol. 70, Print ISBN: 978-3-642-12213-2, Online ISBN: 978-3-642-12214-9, 2010, pp. 1–5, Published by Springer, Berlin, Heidelberg, DOI: 10.1007/978-3-642-12214-9_1, Available: https://link.springer.com/chapter/10.1007/978-3-642-12214-9_1.

[41] Mokhtar Hussein and B. Manjula, "A Hybrid Approach for SVD and Neural Networks Based Robust Image Watermarking", *International Journal of P2P Network Trends and Technology*, Vol. 8, No. 4, 2018, pp. 1–5, Published by Seventh Sense Research Group, DOI: 10.14445/22492615/IJPTT-V8I4P401, Available: https://ijpttjournal.org/archives/ijptt-v8i4p401.

[42] Jiaqi Yang, Xuequan Lu and Wenzhi Chen, "A robust scheme for copy detection of 3D object point clouds", *Neurocomputing*, Print ISSN: 0925-2312, Online ISSN: 1872-8286, Vol. 510, 21 October 2022, pp. 181–192, Published by Elsevier, DOI: 10.1016/j.neucom.2022.09.008, Available: https://www.sciencedirect.com/science/article/abs/pii/S0925231222010967.

[43] Yulan Guo, Hanyun Wang, Qingyong Hu, Hao Liu, Li Liu, *et al.*, "Deep Learning for 3D Point Clouds: A Survey", *IEEE Transactions on Pattern Analysis and Machine Intelligence*, Print ISSN: 0162-8828, Online ISSN: 1939-3539, Vol. 43, No. 12, 1 December 2021, pp. 4338–4364, Published by IEEE, DOI: 10.1109/TPAMI.2020.3005434, Available: https://ieeexplore.ieee.org/document/9127813.

[44] Xin Zhong, Arjon Das, Fahad Alrasheedi and Abdullah Tanvir, "A Brief, In-Depth Survey of Deep Learning-Based Image Watermarking", *Applied Sciences*, Online ISSN: 2076-3417, Vol. 13, No. 21, 30 October 2023, Art. No. 11852, Published by MDPI, DOI: 10.3390/app132111852, Available: https://www.mdpi.com/2076-3417/13/21/11852.